\documentclass[runningheads]{llncs}
\usepackage{graphicx}
\usepackage{array}
\usepackage{float}
\begin{document}

\title{Combination of abstractive and extractive approaches for summarization of long scientific texts}
\titlerunning{Using abstractive and extractive summarization}
%

\author{Tretyak Vladislav \email{191669@niuitmo.ru} \and
Stepanov Denis \email{denis.stepanov@jetbrains.com}}
\authorrunning{V. Tretyak et al.}
%

\institute{ITMO University, \\ 
            49 Kronverksky Pr., St. Peterburg, 197101}

\maketitle              
\begin{abstract}
In this research work, we present a method to generate summaries of long scientific documents that uses the advantages of both extractive and abstractive approaches. Before producing a summary in an abstractive manner, we perform the extractive step, which then is used for conditioning the abstractor module. We used pre-trained transformer-based language models, for both extractor and abstractor. Our experiments showed that using extractive and abstractive models jointly significantly improves summarization results and ROUGE scores.

\end{abstract}
\section{Introduction}
\paragraph{} The language modeling task in the general case is a process of learning the joint probability function of sequences of words in a natural language. Statistical Language Modeling, or Language Modeling, is the development of probabilistic models that are able to predict the next word in the sequence given the words that precede, it also known as context. Language models could operate on different sequence levels. Small language models work with sequences of chars and words. While the big language model works with sentences. But most common language model operates with sequences of words.
\paragraph{} The language model could be used standalone to predict words according to context but usually, the language model is used to solve more challenging tasks. For example, it helps to solve a large range of natural language tasks, such as machine translation, natural language understanding, speech recognition, information retrieval, text summarization. In other words, language models are used in most real-world natural language processing applications and the quality of such applications mostly depends on the language model’s performance. That’s why language modeling task plays a major role in natural language processing and artificial intelligence research.
\paragraph{} The first neural language model used feed-forward architecture. One of the main features of using neural language models is getting representation vector of words sequences. These word vectors usually called embedding vector \cite{mikolov2013distributed} and embeddings for similar words located closer to each other in dimension space, also having similar representations. After successful usage of feed-forward networks, recurrent neural networks \cite{gal2016theoretically} achieved better results in language modeling tasks because of its ability to take into account the position of words in sentences and producing contextual word embeddings \cite{peters2018deep}. Long Short-Term Memory networks \cite{hochreiter1997long} allows the language model to learn the relevant context of longer sequences than feed-forward or RNN's, because of its more sophisticated memory mechanism. Next, the attention mechanism \cite{hochreiter1997long} made improvement in language modeling tasks with a combination of sequence to sequence framework \cite{sutskever2014sequence}. The attention mechanism improves memory mechanism of recurrent neural networks by giving the ability for the decoder network to look at the whole context.
\paragraph{} The next big step in the language modeling task was developing transformer\cite{vaswani2017attention} architecture, with novel self-attention mechanism, that helps the model to use the context of the sentence more efficiently. Such models could take into account both left and right context of the sequence as it is implemented in BERT model \cite{devlin2018bert} and only left context like in GPT model \cite{radford2019language}. Transformer-based LM have own disadvantages, one of them is a limited receptive field. It means that the transformer could only proceed sequences that have limited length, while recurrent neural networks could work with unlimited sequences. This issue partially solved by the Transformer-XL model \cite{dai2019transformer}, which could work with continuous sequences of texts like recurrent neural networks. Despite of this disadvantage, models like GPT with large receptive field and trained on large amount of data are capable of capturing long range dependencies.
\paragraph{} For this work, we propose method that uses both extractive and abstractive approaches for summarization task. Our work improves previous approach \cite{subramanian2019extractive}, by using pre-trained LM instead of training it from scratch. In this research work, we used arxiv dataset as an great example of long scientific documents and in order to compare our work with previous approaches. We split training process in two steps. First, we train extractive model as a classification task, that simply selects which sentences should be selected into summary. Second, we use extracted summary together with different article sections as conditioning for generating abstractive summary. Adding extractive summary into conditioning part is crucial, for generating target summary. Also, we made experiments with different variants of conditioning and found the best combination for it. According to our experiments, extracted summary + introduction and conclusion of the paper performs the best.

\paragraph{}The contributions of this work are:
\begin{itemize}
  \item We show that combination of extractive and abstractive approaches improves quality of produced summary, while using pre-trained transformer-based language models
  \item As a result of applying proposed model, improved ROUGE metric on arxiv dataset
\end{itemize}

\section{Related Work}
Automatic summarization is the process of shortening a set of data computationally, to create a summary that represents the most important or relevant information within the original content. Summarization is one of the main tasks in Natural Language understanding. There are two main approaches for creating a summary. \textbf{Extractive} summarization aims at identifying the information that is then extracted and grouped together to form an extractive summary. It means that, while we train a machine learning model, it solves the classification task. Which information should be included in the summary. \textbf{Abstractive summary} generation rewrites the entire document by building internal semantic representation, and then a summary is created using natural language processing. As an example of the ML model, it receives a source document and generates a summary word by word. Without copying text from source document. The first summarization systems were using extractive techniques, as it is much easier than abstractive. (Hsu et al. 2018) \cite{hsu2018unified} combined both extractive and abstractive approaches for creating summaries. Authors proposed a unified model, that combines sentence-level and word-level attention to take advantage of two approaches. Also, the authors created a novel inconsistency loss function in order to be sure that the proposed model is mutually beneficial to both extractive and abstractive summarization. Another attempt of using both extractive and abstractive models together was proposed in \cite{bae2019summary} paper. The authors proposed a model that uses BERT to perform extractive summarization with the LSTM Pointer network that selects which sentences should be included in the summary. Then, the extracted summary is fed into attention based seq2seq with the copying mechanism for handling out-of-vocabulary (OOV) words. This part of the model is responsible for paraphrasing of the extracted summary and producing an abstractive summary. They used reinforcement learning technique to maximize the ROUGE metric directly instead of maximizing log-likelihood for the next word given the previous ground truth words. Both extractive and abstractive models are trained end2end. In work \cite{liu2019text} authors took pre-trained BERT model and first trained it on extractive task using CNN/Daily Mail datasets. Then, they continue training the same model but doing abstractive summarization.(Subramanian et al. 2019) \cite{subramanian2019extractive} used arxiv dataset for summarizing long scientific texts and uses the similar approach to ours. Their model consists of two parts: first is a hierarchical representation of the document that points out to sentences in the document in order to construct an extractive summary, second is a transformer language model that conditions on the extractive summary, and some part of the paper text as well.
 

\section{Proposed model}
Our model consist of two components. 1) Extractive model classifier that choose which sentences from source text should be included in summary. 2) Abstractive model that uses condition text to produce abstractive summary.
During research work, we propose a model, that combines advantages of extractive and abstractive approaches for creating summaries. The proposed model consists of two parts: the first part is the extractive summarization model, second is the abstractive model. A similar approach was used in \cite{subramanian2019extractive}] paper. The difference is that the extractive and abstractive models are replaced by pre-trained transformer-based LMs.The motivation for this is that pre-trained language model in most cases shows a significant increase in performance. In the previous approach, the transformer model was used as the abstractive model. Instead of it, autoregressive pre-trained LM was used. During research work, a list of experiments for a conditional generation was performed using an extractive summary to generate abstractive summaries. 
The training is performed in two steps. First, the extractive model is trained. Second, training of the abstractive model that uses extractive summary to produce better results

\subsection{Dataset description}
 As it was already mentioned, the dataset from arxiv.org was used for training. This dataset contains long sequences of texts from the scientific domain, with ground truth summaries. The structure of the dataset contains such information as: “article id”, “abstract text”, “article text”, “labels”, “section name”, “sections”. The field “article text” holds full text of a paper, “section name” is a list of paper’s sections, “sections” contains full text of paper divided into sections. That means, we could identify which text is from the introduction section, which is from the abstract section or conclusion. This advantage was used in experiments. The dataset statistics are described in table \ref{tab1}. 
 
\begin{table}
\centering
\caption{Arxiv dataset statistics.}\label{tab1}
\begin{tabular}{| m{7em} | m{7em} | m{7em} |}
\hline
\textbf{num. of docs} &  \textbf{avg. doc. length (words)} & \textbf{avg. summary length (words)}\\
\hline
215.00 &  4938 & 220\\
\hline
\end{tabular}
\end{table}
 
 \paragraph{}Both for validation and test parts we took 5 percent of documents. During the preprocessing step, too long and too short papers were removed, also papers without abstracts, and text of paper were removed. Also, we replaced some LaTeX markup with special tokens such as: [math], [graph], [table], [equation] in order to help model recognize special tokens, other LaTeX source code we cleared. Also, we removed all irrelevant chars and exclude all not Latin letters. This preprocessing pipeline was applied for both extractive and abstractive tasks. We used common approach for creating dataset for training extractive model.
 \paragraph{} First, we create list of sentence pairs, "abstract sentence" $->$ "sentence from paper", every sentence from abstract are matched with every sentence from paper text. Between every pair, we compute the ROUGE metric. In particular, we compute: ROUGE-1, ROUGE-2, ROUGE-L, and take the average value of f-score. After that, we got a scoring value for each pair of sentences, the higher the better. We choose two pairs with the highest scores, this is our positive examples. Then we randomly sample two sentences from the paper text and mark such pairs as negative examples. After completing these steps, we got a dataset that contains a list of sentence pairs and labels. Then, we save the dataset for extractive summarization in a separate file.
 \paragraph{}For the abstractive summarization task, we took our best model that solves extractive summarization tasks and infer it on the dataset, in other words, we generate summaries for each paper in the dataset with our best extractive model. Then we make all necessary preprocessing and save the model generated summaries with corresponding papers and abstracts. These are all steps, that were done with the dataset.

\subsection{Extractive model}
First, we make experiments with extractive models. We used three architectures for extractive summarization models: BERT  \cite{devlin2018bert}, RoBERTa \cite{liu2019roberta}, and ELECTRA \cite{clark2020electra}. ELECTRA has BERT’s architecture but the pre-training phase is different. In the original paper, the authors proposed a new scheme with two models: the generator and language model that are trained jointly. Generator process sentences and corrupts some of the tokens in it. Corruption denotes the process of replacing tokens to some other tokens that could fit into the context of the sentence. In turn, the language model predicts which tokens were replaced by the generator. This scheme of pre-training improves the performance of the language model and reduces the number of training iterations, computing resources. All used extractive models were pre-trained on a large corpus of texts and we took “base” variants, we did not check “large” variants because of long training and hardware constraints. It is highly probable that “large” variants could show even better results. For each of the extractive models, we construct special input of tokens, equation \ref{eq:1}. 

\begin{equation}
 input_{ext} = concat(token_{cls}, seq_{gt}, token_{cls}, seq_{candidate})\label{eq:1}
\end{equation}

\begin{figure}[H]
\centering
\includegraphics[width=0.3\textwidth]{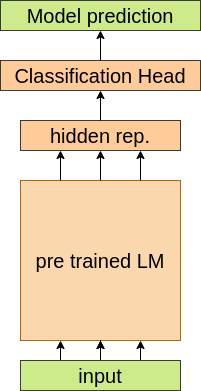}
\caption{Architecture of extractive model}
\label{fig1}
\end{figure}

Where, $x_{ext}$ denotes input to extractive model, $concat()$ denotes concatenation function, $token_{cls}$ denotes special classification token that was originaly used in BERT model, $seq_{gt}$ denotes sentence from ground truth summary, $seq_{candidate}$ corresponds to candidate sentence.
\paragraph{} We used Word Piece tokenizer \cite{schuster2012japanese} for both BERT and ELECTRA models as in original papers, with vocabulary size 30525. RoBERTa uses BPE \cite{sennrich2015neural} tokenizer in original implementation and does not have [CLS] token in its vocabulary, that’s why we manually add it before training. The key benefit of using subword tokenizers is a small dataset size and decreasing out-of-vocabulary cases. Also, special tokens were added: [math], [graph], [table], [equation]. That were extracted by regular expressions while pre-processing. 
\paragraph{}All extractive models were trained using cross-entropy loss as an optimization function. We used the learning rate of  1e-5, model input 512, batch size 24. Also, we applied gradient clipping \cite{pascanu2013difficulty} technique to make the training process more stable. We applied distributed training using two GPUs. The training process took approximately 3-4 days. The architecture of the proposed models is described in Figure \ref{fig1}, below. The scheme remains the same only input to model changes and pre-trained LM  replaced by one of BERT, ELECTRA, RoBERTa. 

\paragraph{}Classification head in the figure above denotes a block that contains a stack of linear layers with an output of size 1 with sigmoid activation. We used ROUGE metrics for evaluation quality of summaries. To evaluate the summarization model, we firstly inference it on the test set. We make a list of pairs: sentence from an abstract sentence from paper and score every pair with the model. Then we use only candidates that have the highest score. After this process, we got the extractive summary that we use with ground truth abstract to calculate ROUGE scores. All scores are averaged between papers. We used ROUGE-1, ROUGE-2, and ROUGE-L as our main evaluation metric for summaries. 
\paragraph{}All proposed models used the same set of training hyperparameters and other settings were the same. From the table above we could conclude that the BERT model achieves the highest result among all other models. With equal set of hyperparameters BERT model achieves better results. In the table above, Oracle denotes the ROUGE score between the ground truth abstract and extracted summary that includes the most relevant sentences from the text of a paper. The Oracle scores in the table indicate the limit for extractive models to get the best summary, according to ROUGE metric. In order to get more coherent text, we perform paraphrasing of extracted summaries. To get paraphrased sentences we apply the back-translation technique. For this, we used pre-trained transformer LM that was trained to translate sentences from english to german and backward. First, we translate the extractive summary into the German language using a pre-trained transformer LM and then back to the English language. Those paraphrased summaries are used later during experiments with condition model. The results of extractive models are presented in table \ref{tab2}.

\begin{table}[H]
\centering
\caption{Extractrive model results on arxiv dataset. Ext denotes extractive.}
\label{tab2}
\begin{tabular}{| m{10em}| m{3em} | m{5em}| m{5em} | m{5em} | }
\hline
\textbf{Model} & \textbf{Type} & \textbf{Rouge-1} & \textbf{Rouge-2} & \textbf{Rouge-L} \\
\hline
BERT & Ext & f1: \textbf{45.4} & f1:  \textbf{20.9} & f1: \textbf{33.7} \\ 
\hline
RoBERTa & Ext & f1: 44.3 & f1:  20.1 & f1: 32.4 \\ 
\hline
ELECTRA & Ext & f1: 41.8 & f1:  17 & f1: 29.3 \\ 
\hline
\end{tabular}
\end{table}

\subsection{Abstractive model}
\paragraph{}After finishing experiments with extractive summarization, we made experiments with abstractive approaches. For abstractive summarization, we used pre-trained autoregressive language models: GPT-2 and BART \cite{lewis2019bart}. We also made experiments with conditional generation for both models. During conditional generation, we give to the model some context, according to which it generates text. We used a condition summary that was extracted by the best extractive model, table \ref{tab2}. Also, we made conditioning on paraphrased summaries. We used an input size of 1024, GPT-base variant. The input is defined by equation \ref{eq:2}. 
\begin{equation}
 input_{abs} = concat(t_{c}, t_{s}) * mask_{segment}\label{eq:2}
\end{equation}
Where we concatenate $t_{c}$ conditioning text with target summary $t_{s}$ and apply a segment mask $mask_{segment}$ that identifies which part of the input is condition and which is target text.
\paragraph{}The experiments with the BART model were performed in a similar way. But there are some differences in model input. BART consists of two parts, the first one is an encoder which uses BERT-like architecture. The second part is a decoder that consists of stacked transformer decoders, similar architecture to GPT. These two parts are connected so that encoder output is fed into the decoder. In such architecture decoder uses hidden states that were produced by encoder. Encoder and decoder are trained end to end. It means that during backward pass, we update weights of decoder and encoder. In such a scenario, we fed the conditioning part into BART’s encoder and target text into the decoder during the training process. We propose different conditioning scenarios. We made conditioning: on the extractive summary, on the introduction of paper, introduction concatenated with conclusion and introduction concatenated with the extractive summary and with the conclusion. We made the assumption that both the introduction and conclusion concentrate the most valuable information for generating summaries. Because usually in introdiction author describes the problem itself, some details about the proposed method, novelty. In conclusion, authors usually make some recap of what was done, conclusion of results. We did not apply conditioning on long texts to GPT-2, because of restrictions in the input size. The condition part plus target text in most cases will not fit into 1024 input size, that's why we only made conditioning on summaries extracted by models. The experiments results with GPT-2 and BART are presented in table \ref{tab3}, below.

\begin{table}
\centering
\caption{Proposed model results on arxiv dataset}
\label{tab3}
\begin{tabular}{ | m{10em}| m{3em} | m{5em}| m{5em} | m{5em} | } 
\hline
\textbf{Model} & \textbf{Type} & \textbf{Rouge-1} & \textbf{Rouge-2} & \textbf{Rouge-L} \\ 
\hline
\multicolumn{5}{|c|}{\textbf{Our approaches}} \\
\hline
BERT & Ext & f1: 45.4 & f1: 20.9 & f1: 33.7 \\ 
\hline
RoBERTa & Ext & f1: 44.3 & f1:  20.1 & f1: 32.4 \\ 
\hline
ELECTRA & Ext & f1: 41.8 & f1:  17 & f1: 29.3 \\ 
\hline
BART conditioned on extractive summary & Mix & f1: 42.5 & f1:  24.6 & f1: 29.2 \\ 
\hline
BART conditioned on introduction & Mix & f1: 40.3 & f1:  21.9 & f1: 27.3 \\
\hline
BART conditioned on introduction with conclusion & Mix & f1: 41.6 & f1:  23.7 & f1: 28.1 \\ 
\hline
BART conditioned on introduction + extractive summary + conclusion & Mix & f1: \textbf{45.3} & f1:  \textbf{25.1} & f1: \textbf{36.2} \\ 
\hline
GPT2 with condition on extractive summary & Mix & f1: 40.1 & f1:  15.8 & f1: 27.7 \\ 
\hline
GPT2 with condition on extractive paraphrased summary  & Mix & f1: 17.9 & f1:  3.5 & f1: 12.2 \\ 
\hline
GPT2 w/o conditioning & Abs & f1: 17.2 & f1: 10.2 & f1: 13.1 \\ 
\hline

\multicolumn{5}{|c|}{\textbf{Previous approaches}} \\
\hline
(Subramanian S. et al. (2019))   & Mix & f1: 42.43 & f1:  15.24 & f1: 24.08 \\ 
\hline
(Cohan A. et al. (2018))  & Mix & f1: 35.80 & f1:  11.05 & f1: 31.8 \\ 
\hline
\multicolumn{5}{|c|}{\textbf{Gold extractive}} \\
\hline
Gold extractive & Oracle & f1: 47.3 & f1:  23.17 & f1: 36 \\ 
\hline
\end{tabular}
\end{table}

\begin{figure}[H]
\centering
\includegraphics[width=0.5\textwidth]{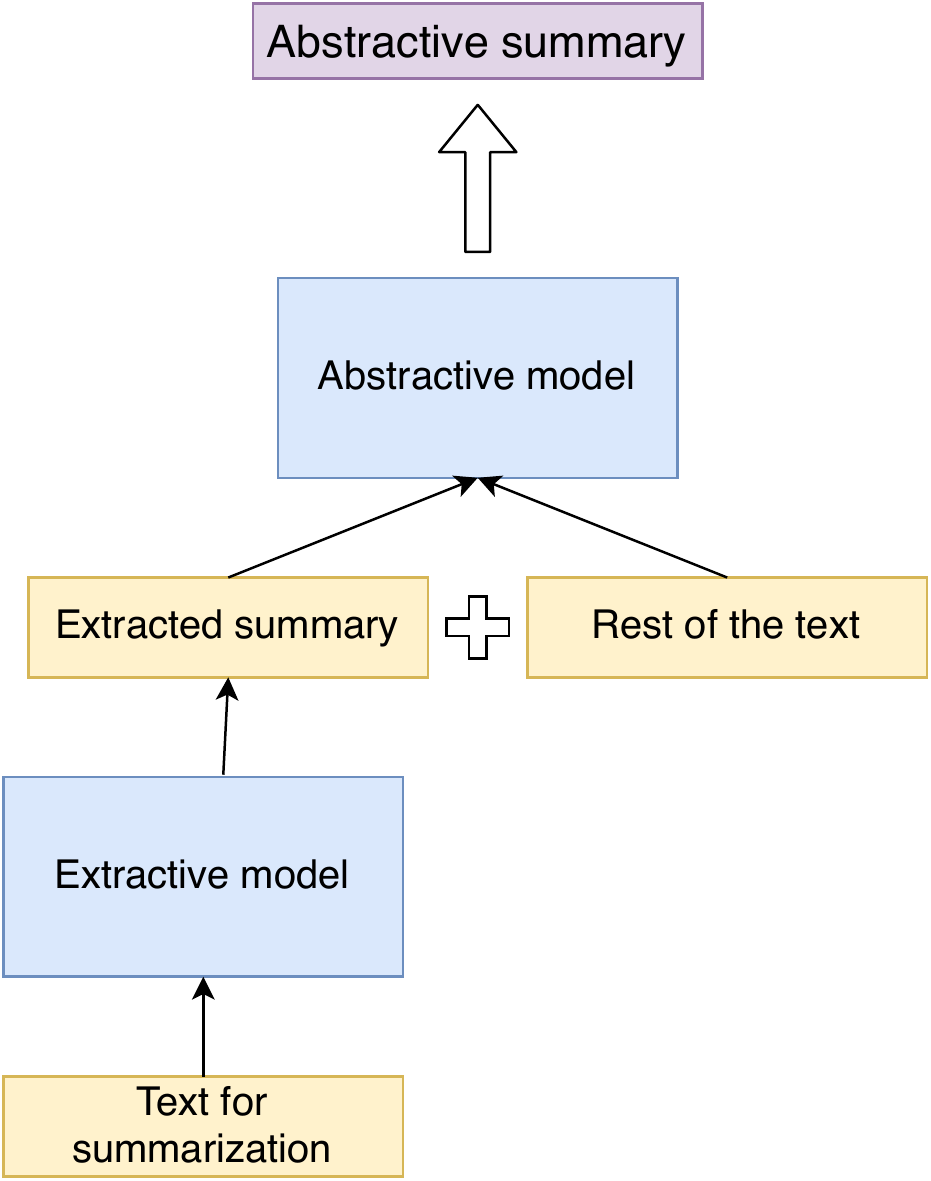}
\caption{Proposed method for generating summaries of long texts}
\label{fig3}
\end{figure}

\paragraph{}From the table above we could conclude that extractive summary plays a crucial role in generating abstractive summaries. Removing the extractive summary from the condition part leads to a decreasing ROUGE score. Also, the assumption that both introduction and conclusion holds the most relevant information, confirmed. It’s obvious that extractive summary has a bigger impact than introduction with conclusion, because extractive summary already holds a lot of relevant (according to ROUGE score) sentences. Also, we investigate that ROUGE-2 and ROUGE-L scores outperforms the best model and outperforms Oracle, because during the abstractive summarization model could produce words that could not be presented in the source document. The best model that uses BERT as extractor and BART as abstractor is presented in Fig. \ref{fig3}. First BERT performs extractive summarization of the article, extracted summary concatenates with the introduction and conclusion of the paper. This setup shows the best performance according to the ROUGE metric and outperforms the previous approach that was applied to arxiv dataset. 

\section{Conclusion}
\paragraph{}The novel improvements was proposed that uses both extractive and abstractive approaches, as an extracted model BERT model was used, as an abstractive model BART model was used. During research work, comparison analysis of different architectures for extractive and abstractive summarization approaches were done. Also, we make experiments with conditioning on different parts of the source document to produce an abstractive summary. Conditioning on extractive summary, introduction and conclusion of the paper shows the best ROUGE score. The assumption about conditioning on paraphrased summary failed, during paraphrasing source sentence is changed so that it becomes irrelevant to ground truth summary. Evaluation results, that were obtained during the research, outperforms previously applied algorithms for arxiv dataset in all ROUGE-1, ROUGE-2 and ROUGE-L metrics \textbf{45.3}, \textbf{25.1} and \textbf{36.2} correspondingly. Also, experiments showed that using the advantages of extractive and abstractive approaches improves the quality of the produced summary, extractive summarization plays a crucial part in generating abstractive summaries.
\paragraph{}As a future improvement of the proposed architecture, end-to-end learning could be applied using both extractor and abstractor. This feature potentially could improve the quality of abstractive summaries. Also, the proposed architecture could be tested for other summarization datasets.

\bibliographystyle{splncs04}
\bibliography{bibliography}

\end{document}